\documentclass[runningheads,a4paper]{llncs}

\usepackage{amssymb}
\usepackage{graphicx}
\usepackage{epstopdf}

\usepackage{url}
\urldef{\mailsa}\path||
\urldef{\mailsb}\path||
\urldef{\mailsc}\path||
\newcommand{\keywords}[1]{\par\addvspace\baselineskip
\noindent\keywordname\enspace\ignorespaces#1}

\begin{document}

\mainmatter  

\title{Automatic Selection of the Optimal Local Feature Detector}

\titlerunning{Automatic Selection of the Optimal Local Feature Detector}

%
%
\author{Bruno Ferrarini\inst{1}%
\and Shoaib Ehsan\inst{1} 
\and Naveed Ur Rehman\inst{2}
\and Ale\u{s} Leonardis\inst{3}
\and Klaus D. McDonald-Maier\inst{1}}

\authorrunning{Automatic Selection of the Optimal Local Feature Detector}

\institute{School of Computer Science and Electronic Engineering, University of Essex, Colchester, UK\\
\email{\{bferra, sehsan, kdm\}@essex.ac.uk}
\and Dep. of Electrical Engineering, COMSATS Institute of Information Technology, Islamabad, Pakistan\\
\email{naveed.rehman@comsats.edu.pk}
\and School of Computer Science, University of Birmingham, Birmingham, UK\\
\email{a.leonardis@cs.bham.ac.uk}
\mailsa
\mailsb
\mailsc
}

%
%

\toctitle{Automatic Selection of the Optimal Local Feature Detector}
\tocauthor{Authors' Instructions}
\maketitle

\begin{abstract}
A large number of different local feature detectors have been proposed in the last few years. However, each feature detector has its own strengths and weaknesses that limit its use to a specific range of applications.
In this paper is presented a tool capable of quickly analysing input images to determine which type and amount of transformation is applied to them and then selecting the optimal feature detector, which is expected to perform the best.
The results show that the performance and the fast execution time render the proposed tool suitable for real-world vision applications.
\end{abstract}

\section{Introduction}
\label{sec_intro}

Local feature detection is an important and challenging task in most vision applications. A large number of different approaches have been proposed so far \cite{tuytelaars_local_2008}. All these techniques present various strengths and weaknesses, which make detectors' performance dependent on the application and, more generally, on the operating conditions, such as the transformation type and amount 
\cite{Mikolajczyk_2005} 
\cite{ehsan2015assessing}. To overcome this problem, 
an obvious solution is to run multiple feature detectors so that the shortcomings of one detector are countered by the strengths of the other detectors. 
However, the computational demand of such an approach can be high and increases with the number of detectors employed. An alternative solution consists of a tool capable of automatically selecting the optimal feature detector to cope with any operating conditions as suggested in \cite{tuytelaars_local_2008}. To the best of our knowledge, such idea has received none or little attention so far \cite{ehsan2013rapid}. This paper aims to bridge this gap by proposing a tool which can determine the transformation type ($T$) and amount ($A$) of input images and then select the detector that is expected to perform the best under those particular operating conditions.
The proposed approach requires to have a prior knowledge of how feature detectors perform under any of the considered operating conditions $(T, A)$. So, in order to design an effective selection stage (Fig. \ref{fig_1}), the evaluation framework proposed in \cite{ferrarini2016performance} is utilized to characterize the performance of a set of feature detectors under varying transformation types and amounts.
This performance characterization, as well as the results presented in this paper, are obtained with the image database available at \cite{ehsan_dataset}. This image database includes 539 scenes, which has been used for generating the datasets for three transformations, namely light reduction, JPEG compression and Gaussian blur. Each dataset has a reference image and several target images, which are obtained by the application of the same transformation to a reference image with increasing amounts. Considering that the JPEG and light reduction datasets include 13 target images and a blur dataset has 9 target images, the resulting number of operating conditions available in the image database \cite{ehsan_dataset} is 18865.\\
The rest of this paper is organized as follows. The proposed selection tool is introduced in Section \ref{sec_sel_tool} while its performance is discussed in Section \ref{sec_results}. Finally, Section \ref{sec_concl}, draws conclusions and discusses the future directions for the automatic selection of the optimal feature detector.

\section{The Automatic Selection Tool}
\label{sec_sel_tool}
\begin{figure}[!t]
   \begin{center}
   \begin{tabular}{c}
   \includegraphics[width=11.3cm, keepaspectratio=true]{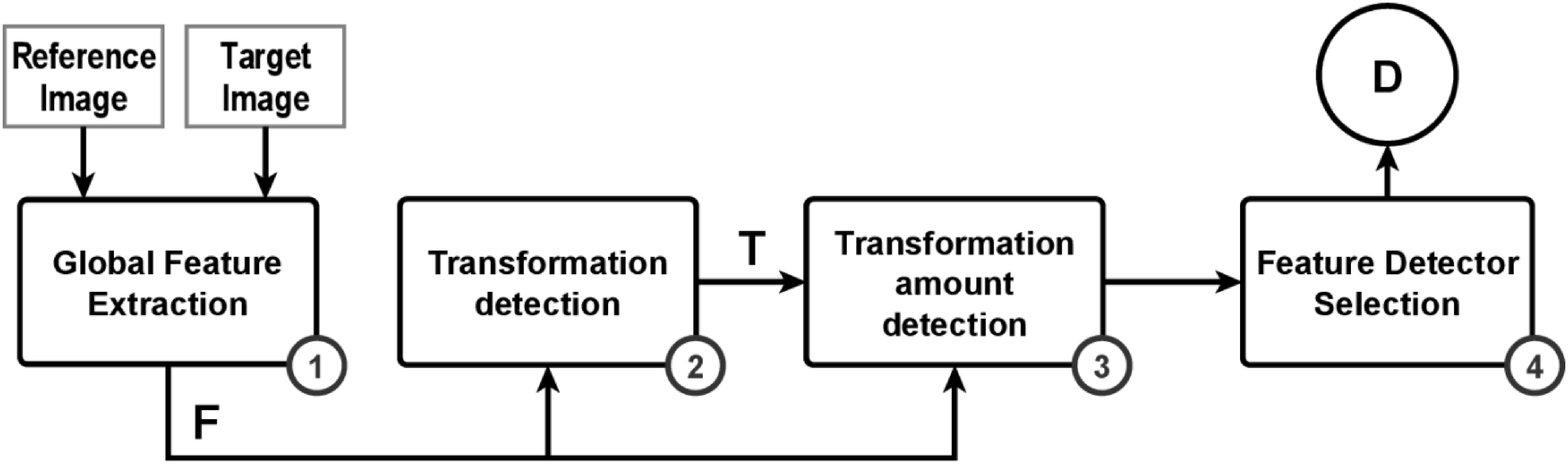}
   \end{tabular}
   \end{center}
   \caption 
   { \label{fig_1} 
The block diagram of the proposed automatic selection system; stage 1 extracts global features from input images, stage 2 and stage 3 determine the operation conditions, whereas the stage 4 selects the optimal feature detector.}
\end{figure}
The proposed system consists of four stages (Fig. \ref{fig_1}). The  first stage extracts global features from the input images, then the second and the third stages determine the type ($T$) and the amount ($A$) of transformation respectively. The last one selects the optimal detector, which is expected to obtain the highest repeatability.  The following subsections describe those four stages of the proposed system and provide more details about the selection criterion of the optimal feature detector.

\subsection{Global Feature Extraction}
\label{ssec_stage1}

The first stage analyses the input pair of target and reference images and then builds a vector of three features: $F = [f_L, f_B, f_L]$. The component $f_L$ is the light reduction feature and is computed as the ratio between the mean of the image histogram of the target and the reference images: $f_L = h_t/h_r$. Hence, lower values of $f_L$ correspond to higher amount of light reduction.
The blur amount of an image is estimated with the perceptual blur metric proposed in \cite{crete2007blur}. The Gaussian Blur feature, $f_B$, is computed as the ratio of the perceptual blur indices of the target and reference images respectively: $f_B = b_t/b_r$. A high value of $f_B$ corresponds to a relatively high level of blurring in the target image.
The JPEG feature $f_J$  is computed with the reference-less quality metric proposed in \cite{wang2002no}, which produces a quality index of an image by combining the blockiness and the zero-crossing rate of the image differential signal along vertical and horizontal lines. Higher the compression rate of a JPEG image, lower is the value of $f_J$.\\
\subsection{Transformation Type Detection Stage}
\label{ssec_stage2}

The transformation ($T$) is determined with a Support Vector Machine (SVM) classifier with a linear kernel function. The SVM has been trained utilizing a portion of the datasets \cite{ehsan_dataset} of 339 scenes chosen randomly. The related datasets for light changes, JPEG compression and Gaussian blur are employed to train the classifier. This results in a training set of 11865 feature vectors (13 x 339 for JPEG compression and light reduction, and 9 x 339 for blurring). \\
The overall accuracy of the prediction is above 99\%. Almost all the classification errors occur between blurred and JPEG compressed images at the lowest amounts of transformation (10-20\% of JPEG compression rate and 0.5-1.0$\sigma$ for Gaussian blur).

\subsection{Transformation Amount Detection Stage}
\label{ssec_stage3}
The third stage is composed of a set of SVMs, each specifically trained to predict the amount $A$ of a single transformation type.  So, once $T$ is determined the corresponding SVM is activated to determine the transformation amount from the feature vector $F$.\\
The overall accuracy for light reduction is close to 100\% while the percentage of transformation amounts correctly classified by the JPEG and blur SVMs are just 75\% and 73\% respectively. However, the results presented in Section \ref{sec_results}, show the relatively low accuracy of the JPEG and blur classifiers do not significantly affect the overall performance of the automatic selection system.

\subsection{Selection of the Optimal Feature Detector}
\label{ssec_stage4}

This stage is implemented as a set of rules, which associate each pair $(T, A)$ with the optimal feature detector $D$ to operate under such type and amount of transformation.
The evaluation framework from \cite{ferrarini2016performance} is utilized to characterize the set of feature detectors available at runtime for selection. Such characterization is carried out following the process described in \cite{ferrarini2016performance} utilizing the training set (Section \ref{ssec_stage2}) of 339 datasets per transformation. First, the improved repeatability rate \cite{Ehsan_2010} is computed for each feature detector using the authors' original programs and the parameters values suggested by them. 
The average of the repeatability rates is computed across all the scene images that are undergone to the same type and amount of transformation. For example, the average repeatability of a detector at 20\% of JPEG compression is obtained as the mean of the repeatability scored with the 339 JPEG images compressed at 20\%.\\
Utilizing the outcomes of the performance characterization, the optimal feature detector for any operating condition is identified utilizing the highest average repeatability as a criterion.
The resulting set of associations, $(T,A) \rightarrow D$, is utilized by the proposed tool at runtime to select of the most suitable feature detector for any given input target image.

\section{Results and Discussion}
\label{sec_results}
\begin{figure}[!t]
   \begin{center}
   \begin{tabular}{c}
   \includegraphics[width=11.3cm, keepaspectratio=true]{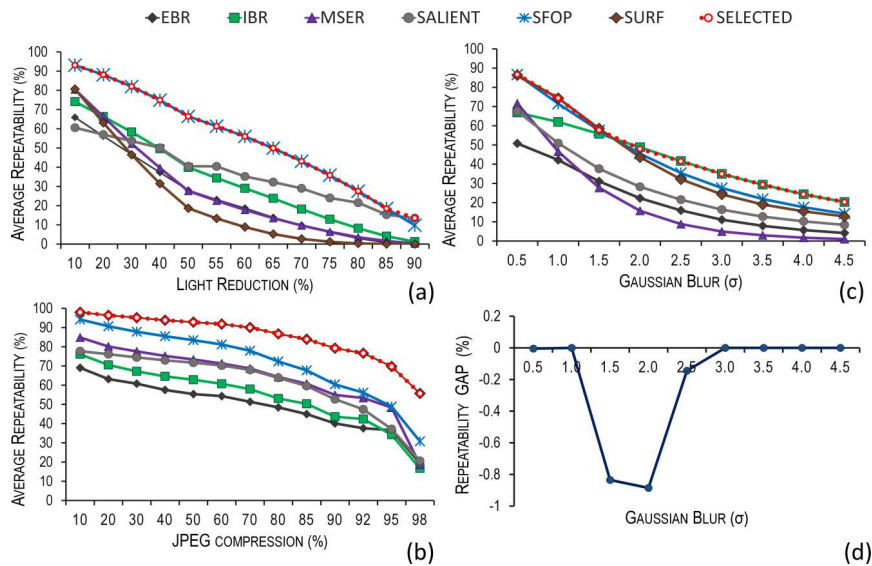}
   \end{tabular}
   \end{center}
   \caption 
   { \label{fig:charts} 
Average repeatability curves of the proposed selection tool and feature detectors working individually for a) light reduction, b) Gaussian blur, c) JPEG compression and d) the average repeatability GAP between the proposed selection tool and feature detectors working individually under Gaussian blur.} 
\end{figure}

This section presents the results of the comparison between the selection algorithm and several feature detectors working individually under varying uniform light reduction, Gaussian blur and JPEG compression. 
The evaluation criteria are the accuracy, which is measured by means of the gap between the average repeatability of the best detector and the optimal detector selected by the tool, and the execution time. 
The employed set of feature detectors represents a variety of different approaches \cite{tuytelaars_local_2008} and includes the following: Edge-Based Region (EBR), Maximally Stable External Region (MSER), Intensity-Based Region (IBR), Salient Regions (SALIENT)  Scale-invariant  Feature  Operator  (SFOP), Speeded Up Robust Features (SURF).
The scenes utilized for the tests are the remaining 200 scenes at \cite{ehsan_dataset}, which are not included in the training set (\ref{ssec_stage2}). Thus, 200 datasets each for light reduction, JPEG compression and blurring transformations have been utilized as a test set. As it is done for characterization of detectors' performance, the repeatability data are obtained using the original authors' programs and with the recommended control parameter values suggested by them.\\
Fig. \ref{fig:charts} shows a comparison of the average repeatability of the feature detectors working individually and the selection algorithm (red dotted line).
Under JPEG compression, the accuracy of the selection is very high with a negligible gap error. Indeed, SURF performs the best under any transformation amount (Fig. \ref{fig:charts}.c), so  the accuracy of the selection depends only on the prediction of the transformation type, which is correct in more then 99\% of the cases. 
The automatic selection tool performs well also with light reduction as it can be appreciated from Fig. \ref{fig:charts}.a where the red dotted line matches perfectly the SFOP's average curve up to 85\% and the SALIENT's curve at 90\% of light reduction (Fig. \ref{fig:charts}.a).\\
To the contrary, under Guassian blur some selection errors occurs as shown in Fig. \ref{fig:charts}.d, where the gap between the average repeatability of the best detector and the one chosen as optimal by the selection tools is plotted. Between 1.5$\sigma$ and 2.0$\sigma$ (Fig. \ref{fig:charts}.d) there is a dip of -1\%. In that range of blurring intensity the average curves of SURF and IBR intersect each other (Fig. \ref{fig:charts}.b) and the wrong predictions of the transformation amount ($A$) causes some errors in the selection of the optimal feature detector. Although, the probability that such classification error occurs is around 9\%, the resulting gap error is just -1\%. This is due to the little difference between the average repeatability values of SURF and IBR, which are close to each other at 1.5$\sigma$ (58.54\% vs 55.78\%) and at 2.0$\sigma$ (43.6\% vs 48.8\%).

A complete run of the proposed tool, from image loading to the detector selection, requires a time comparable to the fastest of the feature detectors considered: MSER. The hardware employed for the test is a laptop equipped with a i7-4710MQ CPU, 16Gb of RAM, and a SATA III SSD Hard drive and the test images have a resolution of 1080 x 717 pixels. MSER and IBR are available as binary executables and have a running time of 150ms and 1.8 seconds respectively while the selection tool, which is a Matlab script, requires 170ms to load images and select a detector. Hence, a system employing the proposed tool with those two feature detectors can extract features in 170 + 150 ms (when MSER is optimal) or 170 ms + 1.8 seconds (when IBR is optimal) while running both MSER and IBR with an image and select the best, would require always more than 1.9 seconds. Thus, the proposed system is equally or more efficient than running more feature detectors with the same image, in addition, it scales really well with the number of feature detectors employed.

\section{Conclusions and Future Directions}
\label{sec_concl}

The automatic tool for selecting the optimal feature detector proposed in this paper represents an attempt to achieve a fully adaptive feature detector system capable of coping with any operating condition. The proposed approach is based on the knowledge of the behaviour of detectors under different operating conditions, which are the transformation type $T$ and the amount of such transformation, $A$. The next step towards a more robust automatic selection system is to consider the scene content as a part of the operating conditions as it is well known that a detector's performance depends also on that factor (\cite{Mikolajczyk_2005}, \cite{ferrarini2015performance}). Thus, modeling the scene content and designing a comprehensive evaluation framework that utilizes it together with the image transformation type and amount should, in our humble opinion, the direction to follow in order to achieve a robust selection tool.

\bibliographystyle{splncs03}

%
%

\end{document}